\pdfoutput=1

\documentclass{ws-ijait-2}

\usepackage[utf8]{inputenc}

\usepackage{combelow} 
\usepackage[hyphens]{url}
\usepackage{mathtools} 

\usepackage{tikz}
\usetikzlibrary{shapes.geometric}
\usepackage{tikz-3dplot}
\usepackage[mode=buildnew]{standalone}
\usepackage{cleveref}
\usepackage{todonotes}

\usepackage{cleveref}

\newcommand{\cmmnt}[1]{\ignorespaces}

\begin{document}

\markboth{Andrei Ionita, André Pomp, Michael Cochez, Tobias Meisen, Stefan Decker}
{Transferring knowledge from monitored to unmonitored areas for forecasting parking spaces}

%
\catchline{}{}{}{}{}
%

\title{TRANSFERRING KNOWLEDGE FROM MONITORED TO UNMONITORED AREAS FOR FORECASTING PARKING SPACES}


\author{ANDREI IONITA}
\address{%
	Computer Science, RWTH Aachen University \\
	Aachen, Germany \\
	andrei.ionita@rwth-aachen.de
}


\author{ANDRÉ POMP}
\address{%
	Institute of Information Management in Mechanical Engineering, RWTH Aachen University,\\
	Aachen Germany \\
	andre.pomp@ima.rwth-aachen.de
}


\author{MICHAEL COCHEZ}
\address{
	Fraunhofer Institute for Applied Information Technology FIT, Aachen, Germany \\
	Department of Computer Science, Vrije Universiteit Amsterdam, Netherlands \\
	Faculty of Information Technology, University of Jyväskylä, Finland \\
	michaelcochez@gmail.com
}

\author{TOBIAS MEISEN}
\address{%
	Chair of Technologies and Management of Digital Transformation, University of Wuppertal, \\
	Wuppertal, Germany \\
	meisen@uni-wuppertal.de
}

\author{STEFAN DECKER}
\address{%
	Computer Science 5, RWTH Aachen University, Germany \\
	Fraunhofer Institute for Applied Information Technology FIT, Aachen, Germany \\
	stefan.decker@dbis.rwth-aachen.de
}

\maketitle

\begin{history}
\begin{center}
	\linespread{0.4} \tiny  \noindent Preprint of an article to be published in Int J. on Artificial Intelligence Tools (IJAIT)  \copyright 2019 [copyright World Scientific Publishing Company] \url{https://www.worldscientific.com/worldscinet/ijait}.
\end{center}
\end{history}

\begin{abstract}
Smart cities around the world have begun monitoring parking areas in order to estimate available parking spots and help drivers looking for parking.
The current results are promising, indeed. 
However, existing approaches are limited by the high cost of sensors that need to be installed throughout the city in order to achieve an accurate estimation.
This work investigates the extension of estimating parking information from areas equipped with sensors to areas where they are missing.
To this end, the similarity between city neighborhoods is determined based on background data, i.e., from geographic information systems.
Using the derived similarity values, we analyze the adaptation of occupancy rates from monitored- to unmonitored parking areas.
\end{abstract}


\pdfoutput=1
\section{Introduction}
	Parking problems and overall traffic congestion are commonplace in cities nowadays. While the number of cars continues to increase\cite{car_increase}, studies show that about 30\% of the traffic in cities is caused by cars that are actively searching for parking\cite{traffic_congestion}. Often it is the lack of planning on behalf of cities that does not accommodate parking facilities proportional to building developments\cite{transport_problems}, which leads to double-parking, more accidents due to distracted drivers, more busy traffic, and, ultimately, a waste of fuel\cite{inrix_study}. As infrastructure solutions are not always optimal and take a long time to implement, there are other strategies to overcome these issues.
	
	Parking can be managed more efficiently if parking spaces would work on an allocation basis, with drivers reserving a spot of their choosing. Accounting for each individual spot and managing its reservations is, however, unrealistic and unsustainable at the moment. Merely providing an overview of areas with free parking spaces would be a big help. Suppose a driver would have access to such a service that indicates free parking spaces at the time when she is arriving in the area. The system would take into account the usual parking levels in the respective area depending on the day of the week and the time of day. Additional information such as current traffic, event data and weather would improve its estimations. Being aware of this sort of parking information beforehand, the driver would pick a traveling path that is less busy through the city and spend significantly less time finding a parking spot that is limited to the area indicated.
	
	This work introduces an approach inspired by the vision to produce parking occupancy estimations for city areas without any previous measurements by taking into account the city infrastructure and the parking data from other areas. This paper extends our previous work\cite{ionita2018park,ionita2017master}, by enriching the evaluation of, and exploring alternative assumptions as, the original approach. The approach distinguishes itself from related research by including open Geographical Information System (GIS) data into the prediction computation.
	
	This paper is organized as follows. Following the introduction, an overview of the research landscape in city parking is presented, before outlining the main assumptions behind our approach. The approach itself is broken down and described in detail, after which its evaluation is set up and carried out. Finally, further possible extensions are outlined and conclusions are drawn.

	\section{Smart Parking Research}
	\label{sec:spresearch}
	Improving the parking situation using sensor data has been subject of research, especially since around 2000. In a 2017 survey\cite{lin,lin2}, Lin compiles an overview of the advances in smart parking by splitting the results into three categories: \textit{information collection}, \textit{system deployment}, and \textit{service dissemination}. We follow their categorization below.
	
	\subsection{Information Collection}
	Under information collection, the survey lists techniques to acquire parking information. Static sensors are usually mounted around parking meters, in the ground or on nearby lamp posts. The dynamic sensors are usually managed through a wireless network infrastructure and are present inside cars, such as taxis, which travel through the city and collect roadside information on parking. In both cases, the transmitted information is the occupancy situation: either the car has left or arrived at a parking space. A number of types of sensors are usually being used in the process: example include infrared sensors, ultrasonic sensors, accelerators, optical sensors, inductive loops, piezoelectric sensors, cameras, and acoustic sensors. Most of the time, the captured data requires post-processing in the form of image or audio recognition before arriving at the target occupancy information. Some captured information also raises privacy issues as it contains sensitive data about the car and driver. Smartphones provide means to collect data and have a great impact through crowdsourcing if the users are given incentives to enable the respective smartphone functions that automatically collect their information. Conversely, parking applications may give drivers incentives to initiate the data transmission themselves and report the parking situation on site. 
	
	Data collection based on smartphone-integrated sensors has been studied in numerous research publications, as it spared the authors to produce and mount special-purpose senors. Xu et al.\cite{xu} makes real-time parking availability estimations based on a system that aggregates the data coming in from mobile phones. The system uses algorithms based on statistical weighted schemes and Kalman filters. Additionally, the authors create parking availability profiles based on historical data and using statistical algorithms. 
	
	Chen et al.\cite{zchen} developed an Android application that finds a parking location at park-and-ride facilities by calculating the probability of parking availability and taking in consideration the shortest travel time. The authors employ fuzzy logic to model the uncertainty of parking availability, with the fuzzy membership function being linear. The authors proposed multiple criteria in finding the best parking location, such as train frequency, service quality, and park-and-ride price. 
	
	PocketParker is a crowdsourcing system, proposed by Nandugudi et al.\cite{nandugudi} that uses smartphone data to predict parking availability. The system is used for parking lots. It requires no input from the user since it notices automatically when a user starts to drive or stops, i.e., departure and arrival events. Based on these, the system builds a probability distribution model that is used to answer queries about parking availability. PocketParker has proved robust to hidden parkers, i.e., parking vehicles that are not using the application. In the authors' simulation, it has reached 94\% rate for parking availability prediction with 105 users over 45 days. 
	
	Koster et al.\cite{koster} propose a smartphone-based solution that recognizes when drivers arrive or leave parking spaces. A Bayesian approach and Hidden Markov Models (HMM) are used to model the parking spaces and respond to user queries for the next parking space. The HMMs are based on gathered historical data. The answer to the user query is a parking space nearby and the probability that it is free at the respective time. The authors emphasize the non-intrusive nature of their solution where drivers only have minimal interaction with their phones to get a recommended free parking space.
	
	\subsection{System Deployment}
	Regarding system deployment, the smart parking survey\cite{lin} refers to the varieties of parking systems, looks into how well they scale, and touches upon the data analysis side. The parking system software is the interface between data sources and the users. Software systems are often in the form of reservation systems, typically run by municipalities or private car parks. These systems may also provide guiding assistance in arriving to the desired parking lot or at the individual parking space. The vacancy prediction component informs the drivers about availability of parking spaces at the destination, either in real-time or at a specific date and time.
	
	There are several publications that investigate parking system worth mentioning. Rajabioun and Ioannou\cite{rajabioun2013} introduce an information system for parking guidance that enables communication between vehicles and the infrastructure. It proposes a prediction algorithm that forecasts the availability of parking locations based on real-time parking information. It takes into account parameters such as parking duration, arrival time, destination, pricing, walking distance, parking capacity, rates of vehicles occupying and leaving parking spots, time restrictions, parking rules, events that disrupt parking availability, etc. Their algorithm uses a probabilistic density distribution model. The parking data was collected both from on-street parking meters and off-street garages in Los Angeles and San Francisco, USA. In a follow-up, Rajabioun and Ioannou\cite{rajabioun2015} propose a multivariate autoregressive model that considers the temporal and spatial correlations of parking availability when making predictions. 
	
	Tiedemann et al.\cite{tiedemann} present the development of a prediction system that estimates occupancy of parking spaces. The occupancy data is collected online via roadside parking sensors and the predictions are realized using neural gas machine learning combined with data threads. The authors notice that some factors play a significant part in the predictions, such as holidays, weather and use the neural gas clustering to separate the data before the data thread method is applied. 
	
	Richter et al.\cite{richter} address the parking prediction problem with the focus on model storage in vehicles. The authors train models of various granularity that would predict parking availability based on the information contained: A one-day model per road segment, a three-day model per road segment, and a seven-day model per road segment. Additionally, models based on regions and time intervals computed by clustering are tried out. Hierarchical clustering with complete linkage is employed. The models are evaluated on street data from the SF\textit{park} project\cite{sfpark_open_data}. The application of clustering before building the models shows a 99\% decrease of model storage space. The prediction success rate is at about 70\%. 
	
	With iParker, Kotb et al.\cite{kotb} propose a system that handles parking reservations. It achieves resource allocation so that drivers pay less for parking, while parking managers receive more resource utilization and hence reach higher revenue. The system is based on mixed-linear programming (MILP). The system uses dynamic resource allocation and pricing models to achieve its goal. In its evaluation, cost cuts for drivers of 28\% were reported, while achieving a 21\% increase in resource utilization and an increased total revenue for parking management by 16\%. 
	
	Shin and Jun\cite{shin} propose an algorithm for smart parking that assigns cars to parking facilities in the city. The criteria based on which the assignment is realized includes driving distance to the parking facility, walking distance from the parking facility to the destination, parking cost, and traffic congestion. The real-time data is collected from parking facilities and from sensors that are integrated in cruising cars. The data is transferred from the central server, where it is managed through a wired/wireless telecommunication network. The authors tested their approach in Luxembourg City. The results of the simulations show improved figures for average driving duration, average walking distance, parking failing rate, parking utilization rate, average standard deviation on the number of guided cars to each parking facility, average occupancy ratio of parking facility, and for the parking facility occupancy rate. 
	
	ParkNet, developed by Mathur et al.\cite{mathur} is a system made up of vehicles that capture parking space information while driving. Every ParkNet vehicle is equipped with a GPS receiver and an ultrasonic sensor facing sideways. The latter determines whether it passes by parking spaces and whether they are occupied. The data is sent to a central server that aggregates it in order to build parking space occupancy maps in real-time. The information is queried by clients that search for a free parking space. The system was evaluated in Highland Park, New Jersey and San Francisco on 500 miles road-side parking data and yield 95\% accurate parking maps and 90\% parking occupancy accuracy. The authors show that the system can further be improved if the sensors are fitted into taxicabs or city buses.
	
	\subsection{Data Dissemination}
	Under data dissemination, the survey\cite{lin} addresses the capability of sharing parking information. This scenario occurs in decentralized parking systems, where the cars find out about free parking spaces in an area where other cars merely drive by and report the real-time situation. 
	
	A selected group of publications is driven by the information exchange approach. Caliskan et al.\cite{caliskan} model the prediction of available parking spaces as a vehicular ad-hoc network (VANET). The network disseminates parking data in order to help the estimation of future occupancy of parking lots. The pieces of disseminated data are timestamp, total capacity of parking lot, number of parking spaces that are currently occupied, the arrival rate, and the parking rate. The latter two are used in the modeling of continuous-time homogeneous Markov chains. The approach is otherwise based on queuing theory. 
	
	Klappenecker et al.\cite{klappenecker} builds on the result of Caliskan and uses an improved version of continuous-time Markov chains for predicting availability of parking spaces. Predictions are communicated between cars in an ad-hoc network. The approach simplifies the computations of transitional probabilities inside a Markov chain model. The system applies to parking lots that are connected to the ad-hoc network. These communicate the number of occupied spaces, capacity, arrival and parking rate. 
	
	Also based on VANETs is Szczurek et al.\cite{szczurek} work, which propose a novel approach that combines machine learning with the information disseminated in ad-hoc vehicular networks. The building blocks of the system are parking reports, which are issued by vehicles leaving a parking space and comprise a report identifier, a location, and a timestamp. The parking reports are being learned by a model, which then indicates whether a parking is available for a specific vehicle. A conditional relevance is used to determine whether a particular report is useful for a specific vehicle. This is modeled using a Naive Bayes method. A parking availability report R is labeled relevant by vehicle V, if the parking space referenced in R is available when V reaches it. Upon evaluation of the methods, the authors reported an improvement in parking discovery times for vehicles.
	
	\section{The SFpark Project}
	To implement and evaluate our proposal, we use the data from the SF\textit{park} project. This project was realized by the San Francisco Municipal Transportation Agency (SFMTA), the city agency that manages the city's transportation, which includes on-street parking\cite{sfpark,sfpark_evaluation}, with the goal to improve parking availability. The SFMTA had the possibility of changing parking rates for on-street parking meters on short notice. Before the project started, parking rates were the same all day, every day, independent of the parking demand. By implementing a demand responsive pricing scheme, parking availability improved dramatically.
	
	Dynamic pricing is a way to control parking occupancy. Parking prices are raised in areas that are almost fully occupied, whereas areas with low parking rates get assigned a lower price. A more advanced version adjusts the prices when enough demands received by the parking system would point to a future parking overload in the respective area.
	
	In conducting the project, nine pilot areas were chosen for monitoring. Out of these areas, seven were selected to have new pricing policies, while two were control areas. The number of metered spaces used was 6,000, which amounts for 25\% of the city's total. The meters allow rates to be deployed remotely, and they transmitted data to a central server through a wireless connection.
	
	The data was collected using parking sensors. These provided the central server with the information needed to calculate the demand-responsive parking rates and provided real-time parking availability information. A parking sensor is a magnetometer that detects changes in the earth's electromagnetic field. A total of 11,700 sensors were deployed, resulting in 8,000 spaces that were equipped with one or two sensors. The sensors delivered valid data from April, 2011 to December, 2013. SF\textit{park} made available real-time information on parking rates and parking occupancy through a smart phone application. The scale and scope of the SF\textit{park} project and its freely available data sets played an important role when choosing to base our project on it.
	
	\section{Rationale Behind Our Approach}
	The approach presented offers a solution to estimating parking occupancy without the help of sensor data. It is based on the observation that parking is determined by the specificities of city areas. Two residential neighborhoods of similar sizes, perhaps far apart from each other, will have very similar parking occupancies: high during nighttime and low during daytime. This will likely differ significantly from office areas, which tend to have most parking spaces occupied during the day and free during the night. Restaurants or shopping centers may represent another distinct category, where customers park usually during the evenings and on weekends, while in the other times they are not very busy, therefore producing a low parking occupancy.
	
	Looking at a city, we can identify a pattern: the types of buildings and the time people spend there determines parking behavior. The presented approach builds on this pattern in order to estimate the level of parking occupancy. Specifically, it uses amenity types and time-spent data to complement established machine learning algorithms in order to arrive at parking levels in places where such forecasts cannot be made only with straightforward models.
	
	The approach does \textit{not} infer parking levels solely based on building metadata and busy times. This information is currently not enough and other factors regarding the city would be needed to arrive at a direct result. Some cities have better parking infrastructures while fitting the same number of people in the offices as cities with scant parking facilities. In the former parking are likely concentrated around the offices, while in the latter the cars are probably distributed uniformly around a larger area around the offices. To circumvent these inconsistencies between cities, we focus on single cities, where parking infrastructure is likely the same given the type of amenities and their dimension. This could be extended to cities in a region or a whole country, depending on the specifics.
	
	The approach therefore uses the (dis)similarities between city areas, with their respective amenity types and time-spent information to help infer parking occupancy. It is assumed that an estimation model can be transferred from a source city area A to a target city area B without any amendment, provided A and B are perfectly similar according to their parking profile. In contrast, the parking occupancy estimation would very much differ if A and B are dissimilar. Specifically, the approach does not offer a precise result in this case, resorting to an interval that expresses the possible parking level.
	
	\subsection{Motivational Example}
	The dimension of the problem we are attempting to solve here is best illustrated with a concrete scenario.
	
	Bob is excited about the interview with a big IT company in his city. He will be driving to the office building located in one of the several office sites in the city. Bob does not like being late, even more so on this occasion, and he wants to leave himself enough buffer time before he arrives at the company reception desk. He has no idea about the parking situation on site, however. In this city, he could spend up to half an hour to find a free parking space. Therefore, Bob uses a new parking app, that can estimate the parking levels at almost every location; the system does not employ sensors everywhere, instead it works by extending the parking behavior from one site to another depending on their specificity, be it offices, restaurants, shopping, or residential. Bob likes the idea and enters his estimated time of arrival at the site and sees that the parking occupancy there will be between 60\% and 80\%. This is good enough for him, he knows that at least 1 out of 5 spaces will be free on average and will likely find a spot in a few minutes. He is suddenly more confident about his punctuality and can now drive more assured to the interview.
	
	\section{Approach}
	The approach, in its generic form, is split into several steps. It begins by acquiring access to data that contains information about parking occupancy for a desired area. The parking data is mapped geographically using OpenStreetMap (OSM).
	Next, the points of interest (POIs) contained in the OSM data are spatially clustered so that the individual clusters are of about the same size. Afterwards, machine learning models are trained on parking data for the computed clusters. For each cluster, mathematical representations are constructed based on the OSM data, which are then used to compute similarity values using \textit{cosine similarity} and \textit{earth mover's distance}. Finally, estimations of parking occupancy are computed by applying the models on areas without parking data with the similarity values factored in.   
	We detail the process step by step below.
	
%
	\subsection{Overview}	

	\begin{romanlist}
		
		\item{\textbf{Get access to appropriate parking data}}
		
		\noindent To have a solid analysis foundation, it is essential to find a well-defined spatial area for which parking measurements over a continuous period of time have been made. Regular status updates, usually by hour, are preferred, if not as soon as they happen. In case multiple distinct data sources for the spatial area and time period are available, limiting oneself to the richest data source is recommended, as multiple sources tend to have different time and space references, and can be inconsistent with regard to sensor errors.
		
		
		\item{\textbf{Map the parking data to OpenStreetMap layers}}
		
		\noindent Another important part is geographically referencing the parking data. OpenStreetMap layers such as points, lines, and polygons that include city artifacts and geographical coordinates together with their metadata are suitable. These are downloaded and associated with the parking occupancy information. The information on amenities included in OSM layers together with the time people spent in amenities is collected as well.
		
		\item{\textbf{Cluster the spatially-referenced data into multiple city areas}}
		
		\noindent Splitting the data corresponding to an entire city into multiple groups is a prerequisite of the approach. Especially, having city areas without parking data completely separated from the city areas with parking data so that the latter can later serve as estimation basis for the former. Splitting is performed spatially. Including any other property, such as OSM metadata, in the clustering algorithm results in noncontinuous areas, which would defeat the purpose of a driver finding a parking space inside a certain radius. Furthermore, the resulting clusters should be of about the same size, as this helps to make inferences later in the process. Averaging the occupancy among the parking spaces inside a cluster, for instance, is less representative for another cluster that has a number of parking spaces of a different order of magnitude.		
		
		\item{\textbf{Build machine learning models for the clustered city areas}}
		
		\noindent A freshly split city area that contains parking data will have a occupancy estimation model. In the training process, the predictor variables includes the measurement timestamp, parking lot capacity, and parking price, while the target variable is the parking occupancy. Methods used for building the models are \textit{decision trees}, \textit{support vector machines}, \textit{multilayer perceptrons}, and \textit{boosted trees}.
		
		\item{\textbf{Build mathematical representations for the city areas}}
		
		\noindent Amenities and time spend information are organized in mathematical objects to reflect the parking demand in their respective city areas. \textit{Cluster vectors}, which serve in \textit{cosine similarity} computation and \textit{cluster Gaussians} that contribute to calculating \textit{earth mover's distance} between city areas are used.

		\item{\textbf{Compute similarity values between any two city areas}}
		
		\noindent The built mathematical representations make it possible to compute similarity measures between city areas. \textit{Cosine similarity} and \textit{earth mover's distance} are defined and computed for every pair of city areas.
		
		\item{\textbf{Apply models on city areas that do not have parking information}}
		
		\noindent When computing the occupancy in clustered city areas with no parking data, the elements built up to now come together. Basically, the machine learning models are applied to the clustered areas without parking data. In the result, the similarity measure between the originating model area and the target area is factored in. In practice, this means that input data for the models will need to be constructed. The result output of the model will be extended in form of an interval upon applying the similarity value: the smaller the similarity, the more the interval will be stretched around the original occupancy result. Occupancy values are expressed between 0\% and 100\%.
		
	\end{romanlist}
	
	\subsection{Getting Access to Parking Data}
	We consider the following types of data as parking data: 
	\begin{romanlist}
		\item \textit{parking occupancy} contains information on the availability of parking spaces at a defined location
		\item \textit{traffic data} contains information regarding the street traffic intensity
		\item \textit{weather data} contains temperature and rain information for the geographical location considered
		\item \textit{event data} contains information relating to events such as street closures which may have an impact on parking
		\item \textit{parking revenue data} contains economic information on parking pricing
		\item \textit{fuel price data} contains prices of fuel in the region
	\end{romanlist}
	
	A detailed overview of the SF\textit{park} dataset is shown in \cref{tab:sfpark_data}.
	
	\begin{table}
		\tbl{Overview of the properties available in the data used from the SF\textit{park} project.}
		{\begin{tabular}{lp{4cm}lp{4cm}}	
				\toprule
				Parking Occupancy & & Traffic & \\
				\colrule
				timestamp & Recorded at full hours & timestamp & Recorded at full hours or in periodic time intervals \\
				parking capacity & The total number of parking spaces at the given location & traffic value & Expressed as average traffic road occupancy, average vehicle count, median speed, or average speed of the traveling cars \\
				parking price & The price of a ticket in dollars at the certain location and the given time & location unit & Given as entire street \\
				parking occupancy & Expressed either as rate (subunitary fraction or percent) or in absolute numbers & & \\
				location unit & Given as street block & & \\
				\colrule
				Events & & Weather & \\
				\colrule
				date and time & Expressed as calendar date or time interval within a day & date & Expressed as calendar date \\
				event name class & Given as the name of the event and its class: road closure or rise of parking demand & temperature & Expressed as the maximum value of the day \\
				location unit & Given as entire street & precipitation & Expressed in the quantity of rain or snow for the corresponding time interval \\
				& & location unit & Given as entire city \\
				\colrule
				Fuel Price & & Parking Revenue & \\
				\colrule
				type of fuel & Provided as gasoline, diesel, etc. & payment type & Expressing the way the driver opted to pay for parking: cash or credit card \\
				price per unit & Provided as the price per gallon & payed amount & Expressed as the amount in US dollars \\
				location unit & Given as entire city & location unit & Given as city district \\
				\botrule
		\end{tabular}}
		\label{tab:sfpark_data}
	\end{table}
		
	Each piece of data is geographically referenced by a \textit{location unit}, i.e., street block, street, district or entire city. In the prospect of using the parking data for training estimation models, the different location unit poses a problem. For the traffic and events data sets, it is entire streets; in the parking revenue dataset, it is city districts, while in case of the weather and fuel price datasets, the location reference is valid for the whole city of San Francisco. Hence, there is a need to align the datasets before they can be used together. In the cases when aggregating values associated to street blocks to the street level is performed, the aggregated values leads to a poorer training performance. Even more so when aggregating street blocks to the city level. Therefore, we were forced to continue without traffic-, events-, parking revenue-, fuel price-, and weather data and rely strictly on the occupancy data further in the process.
	
\begin{figure}[!ht]
		\centering
		\includegraphics[width=0.8\textwidth]{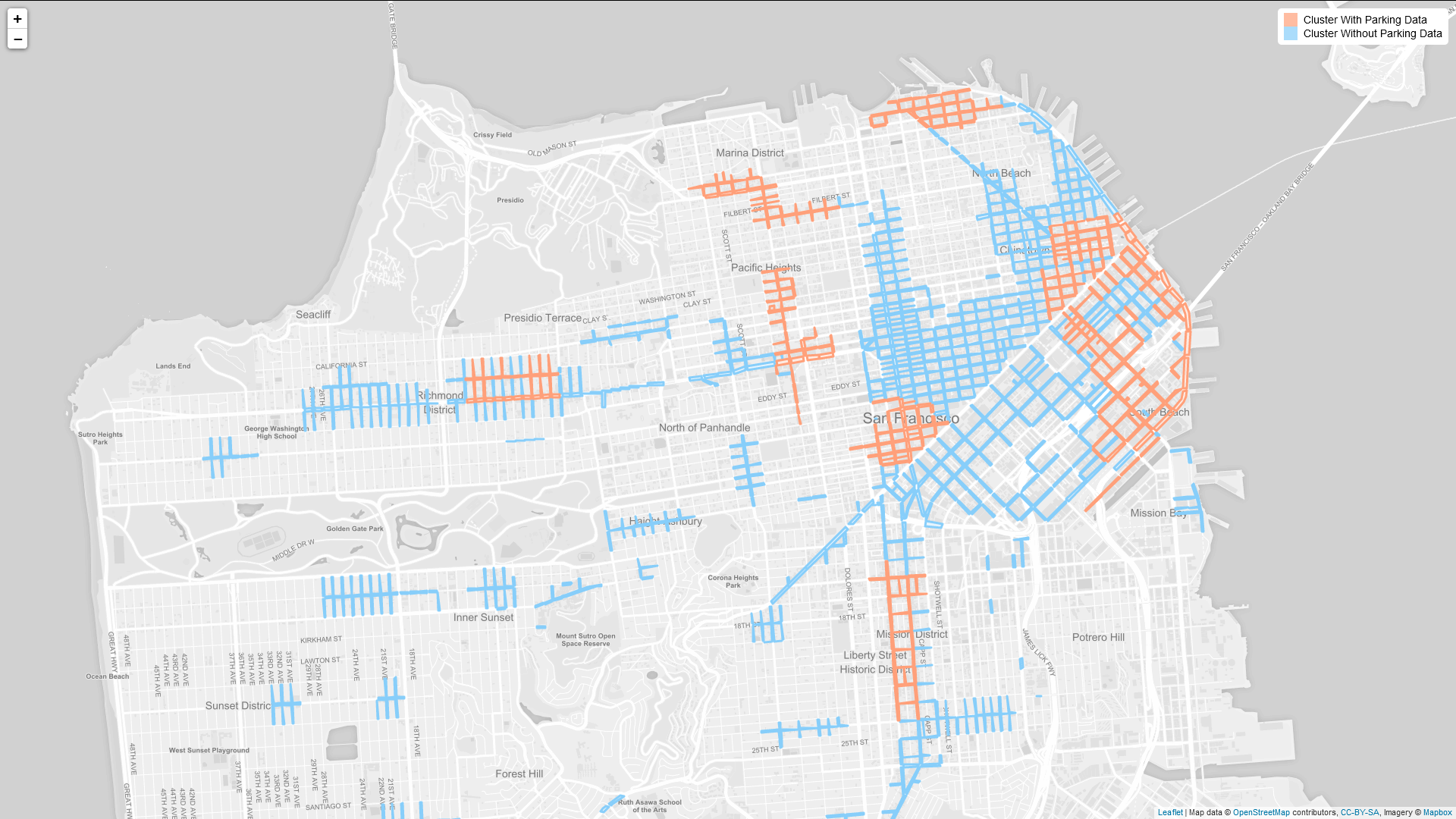}
		\caption{The blocks accounted in SF\textit{park}. The light blue ones are blocks \textit{without} parking data, the light red ones are \textit{with} parking data~\protect\cite{web_application}.}
		\label{fig:before_clustering}
	\end{figure}

	The occupancy data amasses 1.05 million entries with measurements between 04.2011 and 07.2013. The original file provided by SF\textit{park} is about 192M large. The SF\textit{park} data are visualized in \cref{fig:before_clustering} using a Leaflet application built as part of this work.
	
	\subsection{Mapping Parking Data to OpenStreetMap Layers}
\begin{figure}[!ht]
		\centering
		\includegraphics[width=0.8\textwidth]{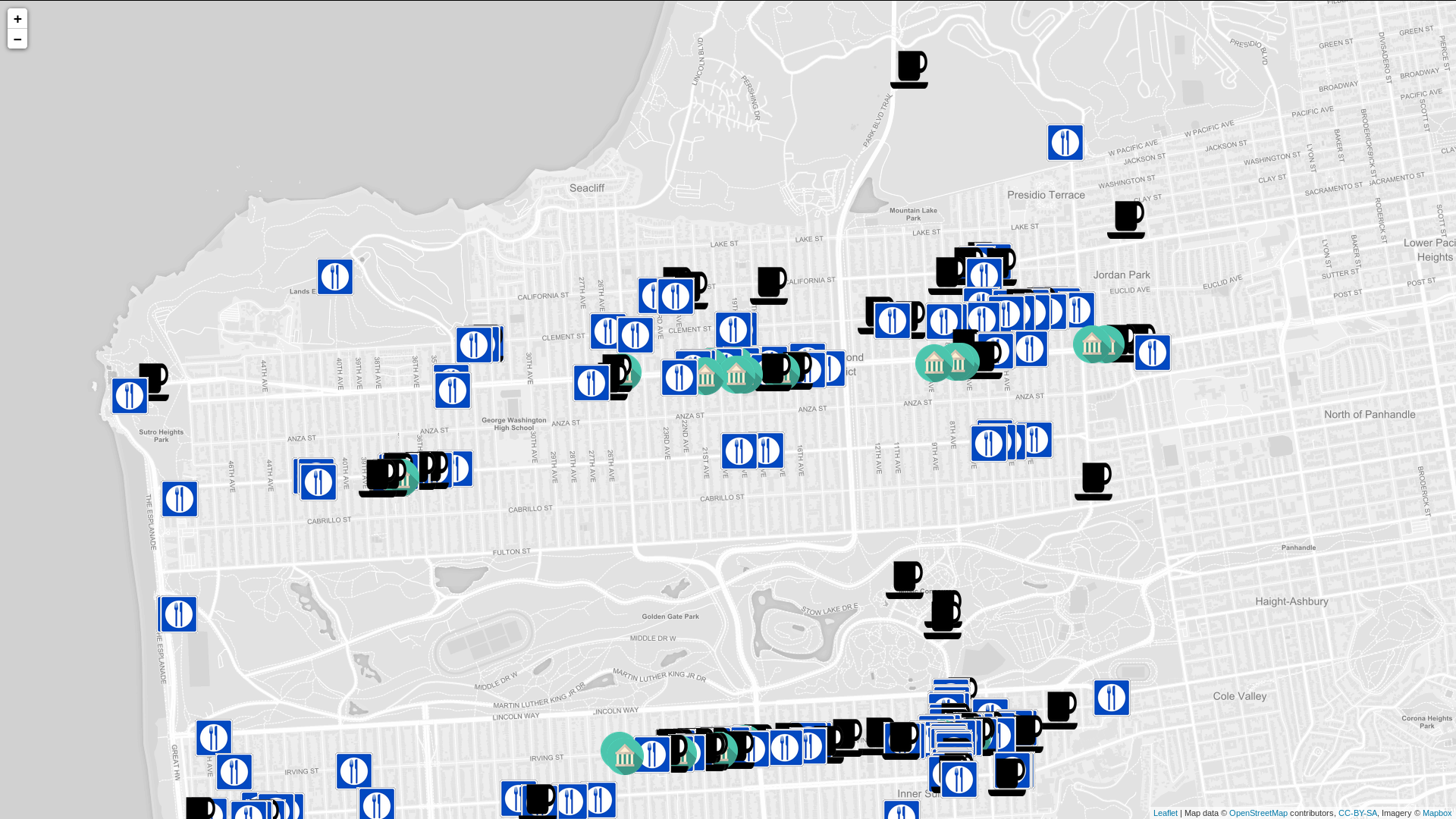}
		\caption{A map indicating public amenities (cafes, restaurants, banks) found at points of interest in OSM (customly built by the authors).}
		\label{fig:pois}
	\end{figure}

	We complement the parking data by downloading OpenSteetMap\footnote{\url{https://www.openstreetmap.org} The maps used in this article are \textcopyright OpenStreetMap contributors.} data corresponding to the location where the parking data belongs to. OSM data is generally available as shapefiles containing the geometry layers: points, polylines, and polygons. We extract the \textit{points of interest} (POIs), which, among multiple attributes, contain the \textit{amenity} attribute indicating the public service, facility, or type of building located at this position as it was annotated by the OSM users (cf. \cref{fig:pois}). The types of the public amenities collected from the POIs are listed in \cref{tab:amenities_list}. The polylines layer contains artifacts mostly in linear form, such as streets or foot paths. Polylines are less interesting for our problem and therefore we ignore them. The polygons layer contains artifacts of polygon shapes such as buildings, parks, university campuses, etc. Polygon objects may contain an amenity attribute as well, in practice the authors have found it often empty, however. When the attribute is present, it enables us to compute the area of the amenity and make an inference towards the capacity of the building.
	
	For the San Francisco area corresponding to the SF\textit{park} project, the OSM file containing the above described layers amounts to about 173M. Inside there are 30,798 POI entries, out of which 5,462 have a non-empty amenity attribute. The number of polygon entries is 147,881.  
	
	\begin{table}[!ht]
		\tbl{List of all OSM amenities found in the SF\textit{park} blocks.}
		{\begin{tabular}{ | l l l l l | }
				\hline
				arts\_centre & dojo & marketplace & shelter & conference\_centre \\
				bank & embassy & music\_rehearsal\_place & shop & fire\_station \\
				bar & fast\_food & music\_school & spa & fuel \\
				biergarten & grocery & nightclub & stripclub & parking \\
				bureau\_de\_change & gym & pet\_grooming\_shop & studio & place\_of\_worship \\
				cafe & hookah\_lounge & pharmacy & training & social\_centre \\
				clinic & ice\_cream & police & veterinary & swimming\_pool \\
				clothes\_store & karaoke & post\_office & vintage\_and\_modern\_resale & theatre \\
				community\_centre & lan\_gaming\_centre & pub & bus\_station & training \\
				dentist & laundry & restaurant & car\_rental & bicycle\_parking \\
				doctors & library & salon & childcare & car\_wash \\
				brokarage & community\_centre & courthouse & fountain & nursing\_home \\
				recycling & social\_facility & toilets & & \\ 
				\hline
		\end{tabular}}
		\label{tab:amenities_list}
	\end{table}
	
\begin{figure}[!ht]
		\centering
		\includegraphics[width=0.8\textwidth]{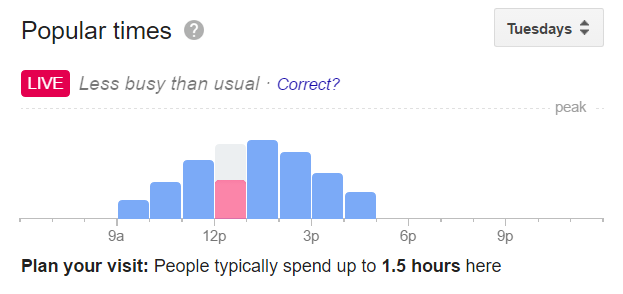}
		\caption{An example of \textit{time spent} information found on Google Places~\protect\cite{google_business}.}
		\label{fig:visit_duration}
	\end{figure}

	Furthermore, we collect the vising duration corresponding to the amenities. We have found that this information is offered by Google Places and FourSquare. The latter data is available via an API, however the access is not free of charge. We used Google Places instead and collected data corresponding to the parking data's location. An example of the service is found within Google Maps for smartphones. It displays typical visiting duration or \textit{time spent} values and popularity of the place for specific time intervals, obtained by Google using a crowsourcing approach that averages the values received from users' smart phone location (cf. \cref{fig:visit_duration}). To obtain the \textit{time spent} values, we manually extract information from 470 places in San Francisco, for which a maximum duration of stay was provided (the minimum duration is not always given)\footnote{This piece of information is not accessible yet via the Google Places API. Google Feature Request: \url{https://issuetracker.google.com/issues/35827350}}. The results are shown in \cref{tab:amenities_google_places} and the \textit{time spent} values are provided in minutes and have been rounded to the nearest integer. We have included only amenities for which at least two stay duration sources were found. 
	
	\begin{table}[!ht]
		\tbl{All amenities listed with their corresponding mean \textit{time spent} information as collected from Google Places. 
			The \emph{cat} column indicates whether the average visiting time is under half an hour (1), 31 to 90 minutes (2), or more than 1.5 hours (3).
		}
		{\begin{tabular}{ | l | c | c | c || l | c | c | c |}
				\hline		
				\textbf{amenity name} & \textbf{mean} & \textbf{stdev} & \textbf{cat} & \textbf{amenity name} & \textbf{mean} & \textbf{stdev} & \textbf{cat} \\ \hline
				arts\_centre  & 110 & 37 & 3 & laundry  & 78 & 16 & 2 \\ \hline
				bank  & 42 & 65 & 2 &  library  & 83 & 13 & 2 \\ \hline
				bar  & 121 & 38 & 3 &  music\_school  & 120 & 30 & 3 \\ \hline
				cafe  & 76 & 39 & 2 &  nightclub  & 189 & 20 & 3 \\ \hline
				clinic  & 100 & 29 & 3 &  pharmacy  & 25 & 20 & 1 \\ \hline
				clothes\_store  & 41 & 37 & 2 &  post\_office  & 16 & 2 & 1 \\ \hline
				community\_centre  & 119 & 40 & 3 &  pub  & 135 & 21 & 3 \\ \hline
				dentist  & 104 & 35 & 3 &  restaurant  & 135 & 32 & 3 \\ \hline
				doctors  & 60 & 42 & 2 &  salon  & 141 & 53 & 3 \\ \hline
				embassy  & 75 & 24 & 2 &  shelter  & 90 & 0 & 2 \\ \hline
				fast\_food  & 31 & 15 & 2 &  shop  & 43 & 21 & 2 \\ \hline
				grocery  & 20 & 10 & 1 &  spa  & 161 & 54 & 3 \\ \hline
				gym  & 100 & 22 & 3 &  stripclub  & 140 & 46 & 3 \\ \hline
				hookah\_lounge  & 130 & 17 & 3 &  studio  & 60 & 0 & 2 \\ \hline
				ice\_cream  & 23 & 7 & 1 &  veterinary  & 67 & 29 & 2 \\ \hline
				karaoke  & 188 & 15 & 3 & {\scriptsize vintage\_modern\_resale}  & 38 & 32 & 2 \\ \hline
		\end{tabular}}
		\label{tab:amenities_google_places}
	\end{table}  
	
	In order to combine the parking and city data, both datasets require a common location unit. For parking occupancy it is street blocks that are provided in latitude and longitude for the coordinate reference system EPSG 4326. The POIs inside the OSM data are expressed in the same geometry reference system and therefore a \textit{merge distance} that matches a parking space to a public amenity can be defined. The \textit{merge distance} can be intuitively understood as the radius around a public amenity. It is defined to represent the parking area that is relevant for a particular public amenity, or, more straightforward, the walking distance from the parked car to e.g., the restaurant, the office, the bank, etc. Based on the above rationale, concrete instances of the merging distance are set to 100m, 200m, and 400m. In \cref{sec:bestmodel}, we discover which one delivers the best results.
	
	\cmmnt{MC: Got till here}
	
	\subsection{The Clustering Process}
	By splitting into city areas, we are making sure that smaller regions lead to more representative parking profiles and therefore parking estimations. As we want an exclusively location-based separation, we may employ K-Means, DBSCAN or OPTICS to cluster the city areas. The distance is calculated between $($latitude$,$ longitude$)$ -pairs of location unit coordinates corresponding to one street block. Since having control over the number of clusters is the goal here, we choose to use K-Means, where we provide the number of expected clusters as input.  In practice, sklearn's $kmeans$ module is used, specifically the K-Means++ algorithm, which initializes the centroids purposefully distant from each other and therefore achieves faster convergence.
	
	There are two clustering processes executed, one for the city area \textit{with} parking data, another one for the city area \textit{without} parking data. The number of clusters chosen in each area is kept proportional to the number of total street blocks that each area contains. It turns out that for SF\textit{park} data the proportion is approximately $2.6$, following the division between the total number of blocks from each group. We have chosen two numbers of clusters to run the evaluation, namely 8 clusters and 16 clusters. The area without parking data will therefore have 20 and 41 clusters, respectively. In the evaluation, we will refer to the number of clusters \textit{with} parking data as \textit{the number of clusters}.	
	
\begin{figure}[!ht]
		\centering
		\includegraphics[width=\textwidth]{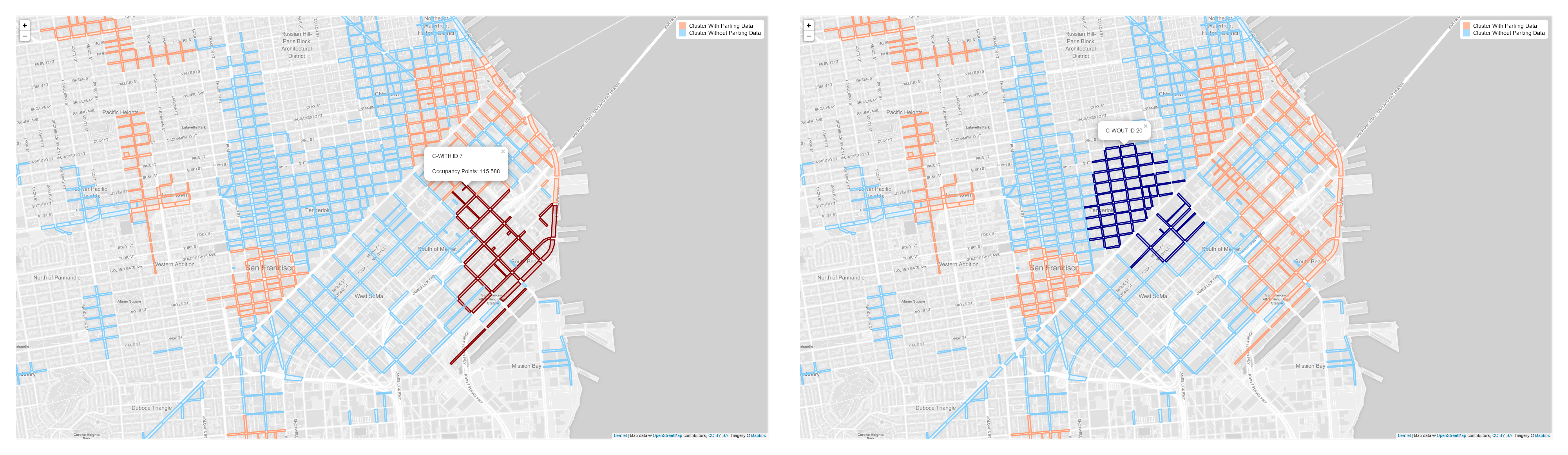}
		\caption{Highlighted cluster \textit{with} parking data on the left side and a cluster \textit{without} parking data on the right side~\protect\cite{web_application}.}
		\label{fig:highlighted_collage}
	\end{figure}

	After running the K-Means clustering process, the Leaflet application map reveals the individual clusters by highlighting them on mouse-over. The clusters \textit{with} parking data will turn dark red, while the clusters \textit{without} parking data will appear in dark blue (cf. \cref{fig:highlighted_collage}). 
	
	\subsection{Building Estimation Models}
	The estimation of parking occupancy is realized using machine learning. We choose to explore this methodology following the solid results machine learning models have delivered for the various smart parking investigations discussed in \cref{sec:spresearch}. A machine learning model $\mathcal{M}$ will be trained for every cluster \textit{with} parking data.
	
	The training data are composed of the parking occupancy data. Specifically, the timestamp, parking capacity and parking price act as predictor variables, while the occupancy rate is the target variable. Before training, data is aggregated across all blocks so that it becomes comparable to other clusters and can be used when training and testing models. The averaging is performed per timestamp, i.e., if multiple blocks have an occupancy record for the same time and block, the occupancy rate will be averaged across both of these. Features such as price and parking capacity per block are averaged as well. See \cref{tab:aggregating_datapoints} for an example of this process. This means that the original collection of data records shrinks, which should decrease the training time. In \cref{tab:models_training_points}, the shrinking rate is shown for various number of clusters. The resulting parking capacity, parking price, and occupancy values are taken as cluster representatives.
	
	\begin{table}
		\tbl{Example of aggregating datapoints}
		{\begin{tabular}{ c }
				\begin{tabular}{ | c | c | c | c | c |}
					\hline
					\textbf{block id} & \textbf{timestamp} & \textbf{price rate} & \textbf{total spots} & \textbf{occupied} \\ \hline
					902   & {2011-04-02 7:00:00} & 0 & 46 & 58 \\ \hline
					32800 & {2011-04-02 7:00:00} & 0 & 32 & 2 \\ \hline
					33005 & {2011-04-02 7:00:00} & 3 & 36 & 12 \\ \hline \hline
					902   & {2011-04-02 8:00:00} & 2 & 46 & 54 \\ \hline
					32800 & {2011-04-02 8:00:00} & 4 & 32 & 5 \\ \hline
					33005 & {2011-04-02 8:00:00} & 3 & 36 & 22 \\ \hline
				\end{tabular} \\
				\colrule
				\begin{tabular}{ | c | c | c | c |}
					\hline
					\textbf{timestamp} & \textbf{price rate} & \textbf{total spots} & \textbf{occupied} \\ \hline
					{2011-04-02 7:00:00} & \textbf{1} & \textbf{38} & \textbf{24} \\ \hline
					{2011-04-02 8:00:00} & \textbf{3} & \textbf{38} & \textbf{27} \\ \hline
				\end{tabular}
		\end{tabular}}
		\begin{tabnote}
			In the first subtable, three distinct blocks belonging to a cluster are transformed into two entries by averaging \textit{price rate}, \textit{total spots} and \textit{occupied} attributes for the two distinct timestamps (second subtable).
		\end{tabnote}
		\label{tab:aggregating_datapoints}
	\end{table}
	
	\begin{table}[!ht]
		\tbl{Number of datapoints aggregated per timestamp vs. all datapoints alongside the shrinking rate for 8, 16 and 32 clusters.}
		{\begin{tabular}{ | c | c | c | c | } 
				\hline
				{cluster size} & {aggregated datapoints} & {all datapoints} & {shrinking rate} \\ \hline
				8 &	9741 & 128525 &	12.3 \\ \hline
				16 & 8409 &	73332 &	8.3 \\ \hline
				32 & 6257 &	29355 &	4.6 \\ \hline
		\end{tabular}}
		\begin{tabnote}
			Values have been averaged across clusters.
		\end{tabnote}
		\label{tab:models_training_points}
	\end{table}
	
	The model training and evaluation is performed in Python via the \textit{scikit-learn} library. During the training phase, we evaluate models such as \textit{decision trees}, \textit{support vector machines}, \textit{multilayer perceptrons} and \textit{boosted trees} using:
	\cmmnt{I would not include these details}
	\begin{enumerate}
		\item $sklearn.tree.DecisionTreeRegressor$,
		\item $sklearn.svm.SVR$,
		\item $sklearn.neural\_network.multilayer\_perceptron.MLPRegressor$, and
		\item $xgboost.XGBRegressor$ respectively.
	\end{enumerate}
	As error metric, we use \textit{root mean square error} (RMSE) and perform a five-fold cross-validation. A model will be evaluated on other clusters with parking occupancy data.
	
\begin{figure}[!ht]
		\centering
		\includegraphics[width=0.8\textwidth]{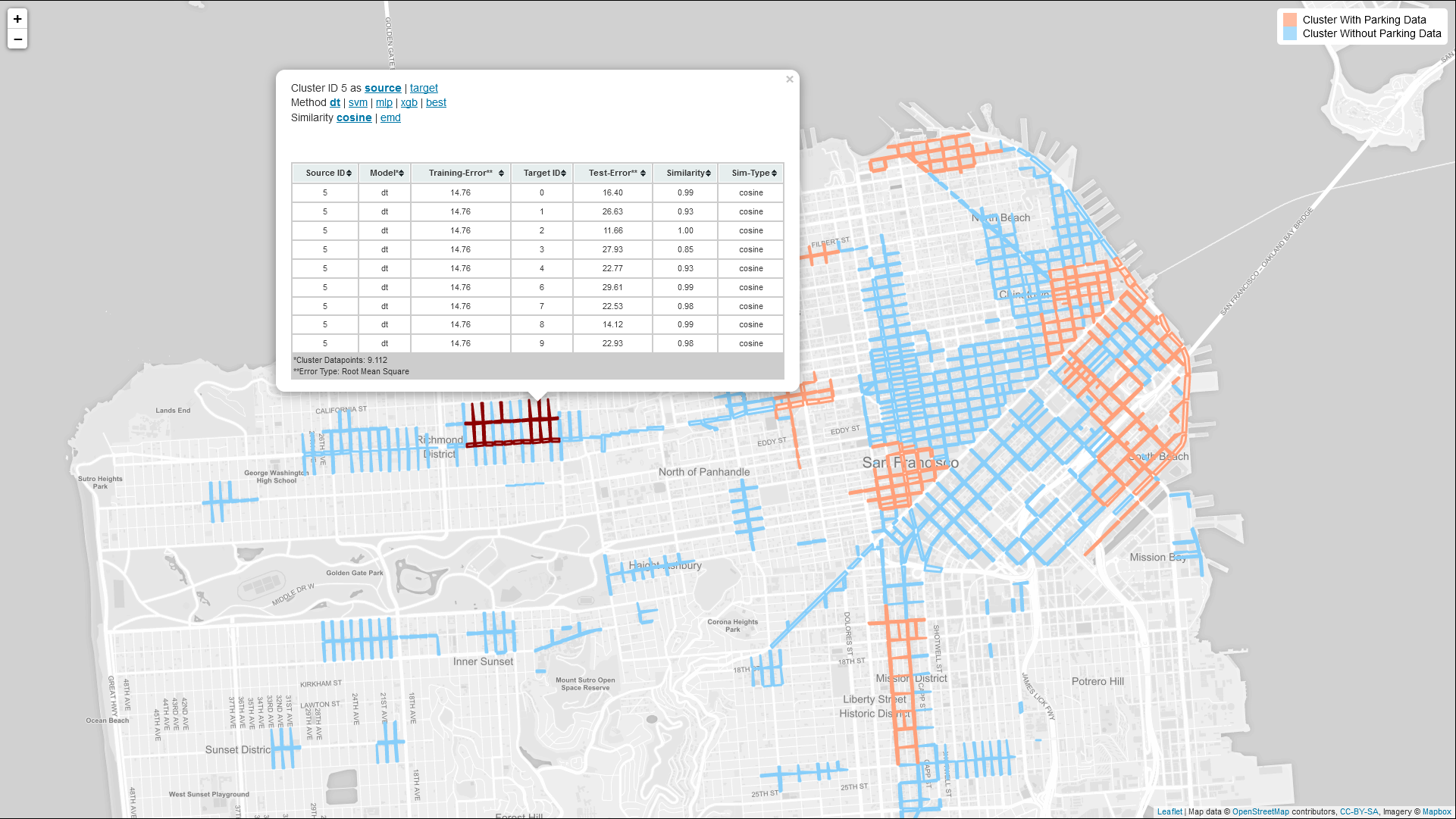}
		\caption{Selected cluster \textit{with} parking data and the pop-up table in the Leaflet application~\protect\cite{web_application}.}
		\label{fig:cwith}
	\end{figure}

\begin{figure}[!ht]
		\centering
		\includegraphics[width=0.8\textwidth]{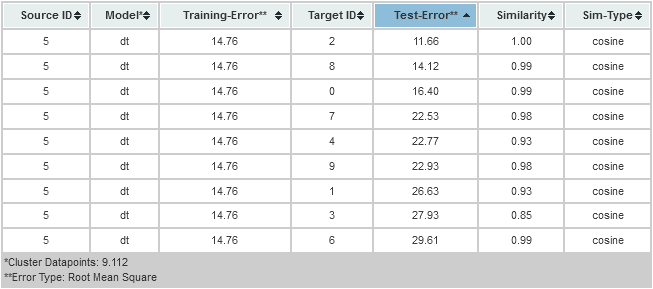}
		\caption{The pop-up table for the Leaflet application view of \cref{fig:cwith}~\protect\cite{web_application}.}
		\label{fig:cwith_table}
	\end{figure}

	Specific details on the actual training of the models can be found in the table \cref{tab:ml_params}. Upon training the models, the clusters can be visualized with the web application as shown in \cref{fig:cwith} as screenshot. \Cref{fig:cwith_table} displays the presented table in more detail.
	
	\begin{table}
		\tbl{Overview of the machine learning methods used, their inputs and parameters.}
		{\begin{tabular}{lp{4cm}lp{4cm}}	
				\toprule
				Predictor Variables & & Target Variable & \\
				\colrule
				timestamp & split into \textit{year} as integer, \textit{calendar week} : [1 - 52] as integer, \textit{weekday} : [1 - 7] as integer, and \textit{hour} : [0 - 23] as integer & parking occupancy & : [0 - 100] as floating point number \\
				parking capacity & as floating point number, when aggregated, otherwise as integer & & \\
				parking price & as floating point number & & \\
				&  & & \\
				\colrule
				Decision Tree Parameters & & SVM Parameters & \\
				\colrule
				Model Selection & randomized search on hyper-parameters for 10 iterations & Model Selection & epsilon-support vector regession model with fixed parameters \\
				$min\_samples\_split$ & : [2, 3, 4, 5] as the minimum number of samples required to split an internal node & $kernel$ & radial basis function \\
				$min\_samples\_leaf$ & : [0.03 - 0.1] as the minimum number of samples required to be a leaf of a node & $C$ & penalty parameters equal to 1 \\
				$max\_features$ & : [0.7, 0.8, 0.9, 1] as the number of features (as fraction from all available features) to consider at each split & $gamma$ & kernel coefficient for the kernel equal to 0.01 \\
				$criterion$ & : [mean squared error, mean absolute error] as the function to measure the quality of a split &  &  \\
				$min\_weight\_fraction\_leaf$ & : [0, 0.1, 0.2] as the minimum weighted fraction of the total sum of weights from all the input samples required to be a leaf node & & \\
				\colrule
				MLP Parameters & & XGB Parameters & \\
				\colrule
				Model Selection & multi-layer perceptron regressor with fixed parameters & Model Selection & exhaustive search over specified parameter values for an extreme gradient boosting model \\
				
				$hidden\_layer\_sizes$ & $(7, 11)$ as tuple representing the number of neurons in each hidden layer & $max\_depth$ & : [2, 3] as maximum tree depth for base learners \\
				
				$max\_iterations$ & 500 as the number of iterations the solver iterates until convergenge & $n\_estimators$ & : [50, 100] as the number of trees to fit\\
				
				& & learning\_rate & : [0.1, 0.25] as boosting learning rate ($\eta$)
				
				\botrule
		\end{tabular}}
		\label{tab:ml_params}
	\end{table}
	
	\subsection{Building Mathematical Representations for City Areas}
	Using amenity data from OSM and time-spent information, we build \textit{cluster vectors} and density estimation kernels.
	
\begin{figure}[!ht]
		\centering
\tdplotsetmaincoords{60}{120}

\newcommand{\Plength}{1}%
\newcommand{\Phoriz}{55}%
\newcommand{\Pvert}{60}%

\begin{tikzpicture}
[
scale=3,
tdplot_main_coords,
axis/.style={->,thick},
vector/.style={-stealth, red, very thick},
vector component/.style={dashed, blue, thick}
]

\draw[axis] (0,0,0) -- (1,0,0) node[anchor=north east] {\footnotesize Cat 1};
\draw[axis] (0,0,0) -- (0,1,0) node[anchor=north west] {\footnotesize Cat 2};
\draw[axis] (0,0,0) -- (0,0,1) node[anchor=south] {\footnotesize Cat 3};

\coordinate (O) at (0,0,0);
\tdplotsetcoord{P}{\Plength}{\Phoriz}{\Pvert}
\draw[vector] (O) -- (P) node[anchor=south west]{\footnotesize Cluster Vector};

\draw[vector component] (O) -- (Pxy);
\draw[vector component] (Pxy) -- (P);

\draw[vector component] (Pxy) -- (Px) node [left]  {};
\draw[vector component] (Pxy) -- (Py) node [above] {};
\draw[vector component] (P) -- (Pz) node [above right] {};

\end{tikzpicture}  
		\caption{An example of a \textit{cluster vector} representing amenity \textit{time spent} information composed from for three categories.}
		\label{fig:cluster_vector}
	\end{figure}
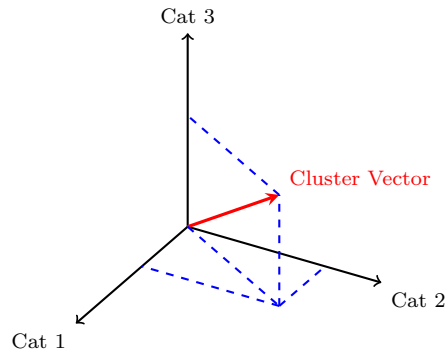

	To form the \textit{cluster vectors}, we first divide all amenities into categories $Cat_1, Cat_2, ..., Cat_n$. The criteria for division will be their average \textit{time spent} values. Each cluster gets represented by an $n$-dimensional vector, whose components correspond to the amenity categories. The magnitude of component $i$ is equal to the number of amenities of category $Cat_i$ that can be found in that particular cluster. For example, a short duration category of up to 30 minutes, a medium duration between 31 and 90 minutes and a large duration of above 90 minutes stay. Compare \cref{fig:cluster_vector} for a general representation.
	
	In the case of San Francisco, the categories are based on the \textit{time spent} mean and are split in three categories. The assigned partitions for every amenity are shown in \cref{tab:amenities_google_places}.
	\begin{romanlist}
		\item $<$ 30 min,
		\item 30 to 90 minutes, and
		\item $>$ 90 minutes. 
	\end{romanlist}
	
\begin{figure}[!ht]
		\centering
		\includegraphics[width=0.7\textwidth]{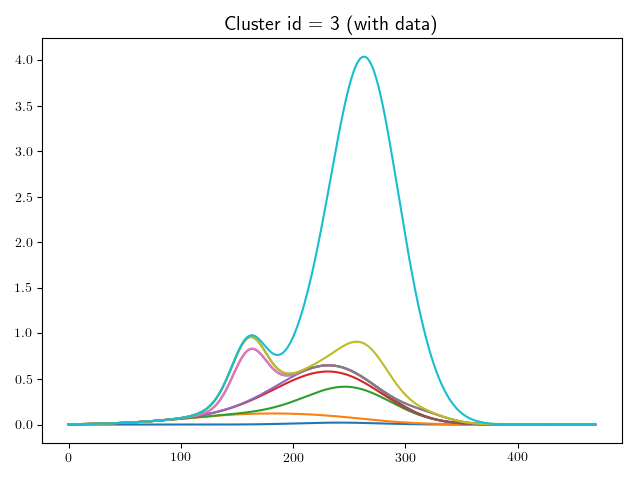}
		\caption{The iterative summing of Gaussian curves representing the amenity \textit{time spent} information resulting in a \textit{cluster Gaussian} as the outer hull (graph obtained inside the Python application using the matplotlib package.}
		\label{fig:gaussian}
	\end{figure}

	A \textit{cluster Gaussian} is a \textit{kernel density estimation} among the probability distribution of an amenity's \textit{time spent} value. We use a Gaussian kernel to express the probability distribution, hence the name. To construct a \textit{cluster Gaussian}, we first collect the mean and standard deviation of the individual amenities' \textit{time spent} values and then we construct its corresponding Gaussian curve. Multiple amenities, each appearing multiple times, will result in a curve that is the linear combination of the individual representations of the amenities as normal distribution curves. Compare \cref{fig:gaussian} for a visualization of the summing process.
	
	\begin{equation}
	emd(\mathcal{C}_i) = \sum_{j=1}^{|amenities|} K_{ij} \times A_j
	\end{equation}
	
	$\forall i \in \{1,..|clusters|\} \text{ and } \forall j \in \{1,..|amenities|\}$
	
	where $A_j$ is an amenity that appears $K_{ij}$ times in the cluster $\mathcal{C}_i$. \\
	
	In practice, the computation is discretized using bins. A bin represents a unit on the $X$ axis, the same on which the \textit{time spent} value is expressed. We will take a number of bins equal to the maximum amenity mean and buffer them with $3\times$ the largest standard deviation, as it is known that within $3\times$ standard deviation on both sides of the mean over 99\% of the Gaussian sum is covered. Moreover, an offset on the $X$-axis equal to $3\times$ the maximum standard deviation is used. This way, we are sure the landscape of summed Gaussians will easily fit into the number of bins. The Gaussians are computed using Python's Optimal Transport package (OT), specifically using the $ot.datasets.make\_1D\_gauss$ module, by providing the number of bins, mean and standard deviation values.
	
	\subsection{Computing Similarities Between City Areas}
	The \textit{cosine similarity} between two vectors is defined as the cosine of the angle between the two vectors. The \textit{cosine similarity} implementation uses the direct mathematical formula by plugging in the magnitudes of the respective vector components.
	
	\begin{equation}
	cos(\theta)=\frac{A\cdot B}{{\lVert}A{\rVert}       
		_2{\lVert}B{\rVert}_2}=\frac{\sum_{i=1}^n{A_iB_i}}{\sqrt{\sum_{i=1}^n{A_i^2}}\sqrt{\sum_{i=1}^n{B_i^2}}}
	\end{equation}
	where $A_i$ and $B_i$ are the components of vector A and respectively. \\
	
	The \textit{earth mover's distance} (emd) is a measure used in statistics that roughly expresses the difference between position and magnitude of two curves. It is best explained by regarding the curves as the hull of earth piles. For two separate earth piles, emd computes the minimum effort of rearranging a pile so that the shape of the other pile is obtained. Moving P particles over a distance D is equal to the effort $P \times D$. A prerequisite for this operation is that the two piles need to contain the same quantity of earth.
	
	More rigorously, emd is better known in mathematics as Wasserstein Metric. Given two normal distributions $\mu_1=\mathcal{N}(m_1,C_1)$ and $\mu_2=\mathcal{N}(m_2,C_2)$, where $m_1$ and $m_2 \in \mathbb{R}^{n}$ are their respective expected values and $C_1$ and $C_2 \in \mathbb{R}^{n\times n}$, their 2-Wasserstein distance between $\mu_1$ and $\mu_2$ is:
	
	\begin{equation}
	W_2(\mu_1,\mu_2)^2={\lVert}m_1-m_2{\rVert}^2_2+trace(C_1+C_2-2(C_2^{1/2}C_1C_2^{1/2})^{1/2})
	\end{equation}
	
	Notice that emd is applicable only when the sum under both Gaussian curves is equal. Therefore, all \textit{cluster Gaussians} will get normalized before emd is computed. In practive, after summing up the respective Gaussians and arriving at the \textit{cluster Gaussians}, the $scipy.stats$ module is used to compute emd by means of the $wasserstein\_distance$ method. Provided here are the bin ranges on the X axis and their Y axis values corresponding to the computed \textit{cluster Gaussians}.
	
	\subsection{Computing Parking Occupancy Estimations}
	Once all models $\mathcal{M}$ have been built for the clusters \textit{with} parking data, making estimations on parking occupancy in these areas is straightforward. The input data fed to the model is manufactured from averaging the predicted variables from the areas with parking data. However, we want to apply these models on the clusters that are missing parking data. We derive the \textit{estimation interval} for cluster $\mathcal{C}_{wout}^j$ based on the model of cluster $\mathcal{C}_{with}^i$ as follows.
	
	For \textit{cosine similarity}:
	\begin{equation}
	E(\mathcal{C}_{with}^i,\mathcal{C}_{wout}^j) = [\mathcal{M}(\mathcal{C}_{with}^i) - (1 - sim_{ij}), \text{    } \mathcal{M}(\mathcal{C}_{with}^i) + (1 - sim_{ij})]
	\end{equation}
	$$\text{where } sim_{ij} = sim(\mathcal{C}_{with}^i,\mathcal{C}_{wout}^j) \in [0,1]$$
	
	For emd:
	\begin{equation}
	E(\mathcal{C}_{with}^i,\mathcal{C}_{wout}^j) = [\mathcal{M}(\mathcal{C}_{with}^i) - emd_{ij},   \text{    }\mathcal{M}(\mathcal{C}_{with}^i) + emd_{ij}]
	\end{equation}
	$$\text{where } emd_{ij} = emd(\mathcal{C}_{with}^i,\mathcal{C}_{wout}^j) \in [0,1]$$
	
	$$\forall i \in \{0,...,|\mathcal{C}_{with}|-1\} \text{ and } \forall j \in \{0,...,|\mathcal{C}_{wout}|-1\}$$.
	
	The result is an \textit{estimation interval} that \textit{stretches} the punctual estimation into an interval depending on the similarity value. The lower the similarity value is, the larger the length of the resulting estimation interval will be.
	
	
	Furthermore, we define an \textit{estimation intersection interval}, whose purpose is to narrow down the computed estimation interval.
	An estimation intersection interval for the clusters $\mathcal{C}_{with}^i$ and $\mathcal{C}_{wout}^j$ is computed by intersecting the \textit{estimation intervals} that have a better similarity among the clusters with data $\mathcal{C}_{with}^{0}, ..., \mathcal{C}_{with}^{i-1}$ and the same cluster without data $\mathcal{C}_{wout}^j$.
	
	\begin{equation}
	EII(\mathcal{C}_{with}^i,\mathcal{C}_{wout}^j) = \bigcap_{k=0}^{i-1} EI(\mathcal{C}_{with}^k,\mathcal{C}_{wout}^j) \\
	\end{equation}
	
	where
	\begin{equation}
	sim(\mathcal{C}_{with}^k,\mathcal{C}_{wout}^j) < sim(\mathcal{C}_{with}^i,\mathcal{C}_{wout}^j), k \in \{0,..,i - 1\} \text{ for emd}\\
	\end{equation}
	\begin{equation}
	sim(\mathcal{C}_{with}^k,\mathcal{C}_{wout}^j) > sim(\mathcal{C}_{with}^i,\mathcal{C}_{wout}^j), k \in \{0,..,i - 1\} \text{ for cosine} 
	\end{equation}
	
	$$\forall i \in \{0,...,|\mathcal{C}_{with}|-1\} \text{ and } \forall j \in \{0,...,|\mathcal{C}_{wout}|-1\}$$.
		
	\section{Evaluation}
	We evaluate various pieces of the system that has been presented. Firstly, we  establish the machine learning method that achieves best results on average across clusters. Afterwards, we compare the cluster models' test errors with the independently-computed similarity values between clusters. More specifically, a \textit{source} cluster's model will tested on a \textit{target} cluster and the error is correlated to the similarity between the \textit{source} and the \textit{target} clusters. Both \textit{cosine} and \textit{emd} functions will be used. The correlations will be expressed as Pearson- and Spearman's rank coefficients. Afterwards, we take a look at the results of applying the models to clusters \textit{without parking data} and visualize the results. 
	
	Furthermore, some alternative method are investigated. Firstly, we look at the model test errors and correlation results by skipping the aggregating step, i.e., instead of averaging the datapoints over timestamp per cluster, we build the cluster models using the entire occupancy data directly. Secondly, we use the \textit{amenity area} as the basis for the similarity functions in calculating the correlations between model test errors and similarity values. Finally, we question whether the similarity function approach is the most efficient and transfer its purpose to the machine learning phrase. The model will receive absolute \textit{cosine vectors} and \textit{cluster Gaussian} values as additional features and its model test error and correlation values will be compared to the ones from the original approach.
	
	\subsection{Best model method}
	\label{sec:bestmodel}
	Of the models were trained using the four methods (\textit{decision trees}, \textit{support vector machines}, \textit{multilayer perceptrons}, and \textit{gradient boosted trees}), \cref{evaluation:best_model_method} shows the distribution of best machine learning methods in case of 8 and 16 clusters. The values were obtained by summing up the number of times a method produced the smallest estimation error, i.e., RMSE, among the four methods for all combinations of clusters with parking data $(\mathcal{C}_{source}, \mathcal{C}_{target})$.
	Extreme gradient boosting claims the first spot in both cases.
	In further experiments in the evaluation we shall only report on models trained using extreme gradient boosting.
	
	\begin{table}[!ht]
		\tbl{The fraction of best models among \textit{decision trees}, \textit{support vector machines}, \textit{multilayer perceptrons} and \textit{extreme gradient boosting} measured as RMSE when applied on all pairs of clusters.}
		{\begin{tabular}{ | c | c | c | c | c |}
				\hline
				& \textbf{dt} & \textbf{svm} & \textbf{mlp} & \textbf{xgb} \\ \hline
				\textbf{8 clusters} & 24.6\% & 17.5\% & 12.3\% & \textbf{45.6\%} \\ \hline
				\textbf{16 clusters} & 14.6\% & 13.8\% & 13.8\% & \textbf{57.9\%} \\ \hline
			\end{tabular}}
			\label{evaluation:best_model_method}
	\end{table}
		
	\subsection{Similarity Values vs. Estimation Errors}
	The central goal of this work is estimate the occupancy of clusters where no parking data is available by using model predictions and pair-wise cluster similarity values. For the purpose of evaluating the models, we shall use clusters with parking data that were left out from the training dataset as the application target of the prediction models, while the computed similarity values will serve to confirm the prediction errors. The higher the absolute correlation between the cluster similarity values and the model test errors, the better the accuracy of the cluster similarity. 
	
	
	 
	Concretely, for every pair of clusters ($\mathcal{C}_{source}, \mathcal{C}_{target}$), a model $\mathcal{M}_{\mathcal{C}_{source}}$ is trained on $\mathcal{C}_{source}$ and its test error ($\mathcal{M}_{\mathcal{C}_{source}}(\mathcal{C}_{target})$) shall be correlated with the cosine and emd similarity values between $\mathcal{C}_{source}$ and $\mathcal{C}_{target}$. Here, we use two correlation coefficients: the Pearson correlation coefficient and Spearman's rank correlation coefficient, which results in four correlation measures: 
	\begin{romanlist}
		\item	cosine (Pearson) correlation
		\item 	cosine (Spearman's) rank correlation
		\item 	emd (Pearson) correlation
		\item 	emd (Spearman's) rank correlation. 
	\end{romanlist}
	The evaluation was performed for configurations of 8 and 16 clusters respectively. Additionally, we varied the \textit{merge distance} between 100m, 200m, and 400m to see how the correlation behaves. In \cref{tab:similarity_vs_estimation} the final results are shown. A correlation result is averaged across all clusters. Due to their mathematical meaning, the \textit{cosine similarity} values below zero express a positive correlation, whereas emd values above zero express a positive correlation. 
	
	We notice that the \textit{cosine similarity} achieves better results than emd for the same testing configuration, peaking for 8 clusters and 100m \textit{merge distance}. Its average Pearson coefficient is $-0.55$, while the mean Spearman rank coefficient is $-0.49$. emd positively correlates the most for the same testing configuration (8 clusters, 100m), when the average Pearson coefficient is at $0.28$ and Spearman's rank coefficient equals $0.23$. There is a clear descending trend in correlations, as the \textit{merge distance} increases.  Also, the results for 8 clusters are superior to the ones when the city is split in 16 clusters. Further in the evaluation runs we fix the \textit{merge distance} to 100m.
	
	\begin{table}[!ht]
		\tbl{Correlations between similarity values and model estimations errors for pairs of clusters \textit{with} parking data ($\mathcal{C}_{source}, \mathcal{C}_{target})$.}
		{\begin{tabular}{ | c | c | c | c | c | c | c | c | c |}
				\hline
				{} & \multicolumn{4}{c|}{8 clusters} \\ \hline
				{merge distance} & cosine & rank\_cosine & emd & rank\_emd \\ \hline
				100m & \textbf{-0.55} & \textbf{-0.49} & \textbf{0.28} & 0.23 \\ \hline
				200m & -0.34 & -0.30 & 0.26	& 0.23 \\ \hline
				400m & -0.23 & -0.08 & 0.25 & \textbf{0.27} \\ \hline
		\end{tabular}}
		\vspace{3em}
		{\begin{tabular}{ | c | c | c | c | c | }
				\hline
				{} & \multicolumn{4}{c|}{16 clusters} \\ \hline
				{merge distance} & cosine & rank\_cosine & emd & rank\_emd \\ \hline
				100m & \textbf{-0.20} & \textbf{-0.17} & \textbf{0.10} & \textbf{0.11} \\ \hline
				200m & -0.13 & -0.11 & 0.02 & 0.02 \\ \hline
				400m & -0.17 & -0.17 & 0.08 & 0.11 \\ \hline
		\end{tabular}}
		\label{tab:similarity_vs_estimation}
		\begin{tabnote}
			For \textit{cosine similarity} the values show a negative correlation tendency, while for the correlation based on emd similarity expresses a positive correlation tendency. The correlations are measured using Pearson coefficient and Spearman's rank coefficient.
		\end{tabnote}
	\end{table}
				
	\subsection{Estimations for clusters without parking data}
\begin{figure}[!ht]
		\centering
		\includegraphics[width=0.8\textwidth]{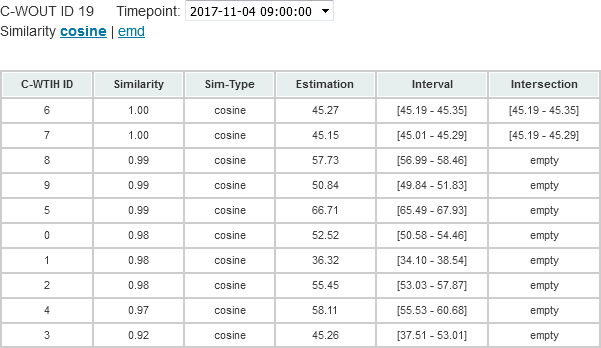}
		\caption{The pop-up table of a cluster without data.
			Notice the drop-down list from which the time can be selected. There are 8 times to select, equally spaced throughout the day that followed our experiment. The similarity values are rounded off. The results are expressed in intervals of occupancy. The successive intersection of intervals succeeds only for the first two values, afterwards it is empty ~\protect\cite{web_application}.}
		\label{fig:cwout_table}
	\end{figure}

	We apply the models trained on SF\textit{park} data on clusters \textit{without} parking data.
	The testing data records are composed of values equal to the averages of the respective data types in the clusters \textit{with} parking data.
	This is the case for \textit{parking price} and \textit{parking capacity}.
	One piece of data that still needs to be provided so that the estimation is computed is the timestamp.
	For convenience, we choose the next day relative to when we ran the expriment and 8 times spread throughout the day.
	A sample of the input values fed to the model is show in \cref{tab:ml_cwout}, while the results of the estimation which includes the similarity values is visualized in the web application as in \cref{fig:cwout_table}.
	
	\begin{table}
		\tbl{Input data for models when predicting occupancy for city areas without parking data. The date has been chosen arbitrarily with several times equally spaced throughout the day. The price rate and total spots values are equal to the approximate averages of the entire parking dataset.}
		{\begin{tabular}{cccc}	
				\toprule
				Date & Time & Price Rate & Total Spots \\
				\colrule
				2017-11-04 & 	00:00 & 1.0 & 20 \\
				& 	03:00 & 1.0 & 20 \\
				& 	06:00 & 1.0 & 20 \\
				& 	09:00 & 1.0 & 20 \\
				& 	12:00 & 1.0 & 20 \\
				& 	15:00 & 1.0 & 20 \\
				& 	18:00 & 1.0 & 20 \\
				& 	21:00 & 1.0 & 20 \\
				\botrule
		\end{tabular}}
		\label{tab:ml_cwout}
	\end{table}
				
	\subsection{Models built on all occupancy datapoints}
	Up to now, the evaluation involved models trained and tested- on aggregated datapoints. 
	We ask ourselves, however, whether a model trained on all datapoints performs better than when tested on an aggregated cluster. Or whether an aggregate model delivers better results on an all-datapoints cluster than a model trained on all datapoints? Here, we experiment these combinations by training models on both all- and aggregate datapoints and apply them on both types of aggregation forms. 
	
	See \cref{tab:train_test_errors} for an overview of test errors for 8 and 16 clusters. One observes that the errors from models applied on aggregated datapoints are about 5 unit points (5\%) smaller than the errors from models applied on all datapoints. This is naturally accounted for by the smaller spread of occupancy values that the aggregation brings with itself. As far as source models are concerned, there is no significant difference between the aggregate and all datapoint models, i.e., the margin is under 1\%. Regarding the number of clusters, the values for 8-cluster models are slightly better than 16-cluster models for aggregated datapoints, but lose when the testing bed is equal to all datapoints.
	
	In \cref{tab:correlation_values} the resulting correlations of testing errors with cosine- and emd similarity values are listed for models of 8 and 16 cluster configurations. It follows that the cosine- and rank cosine correlation values are on average closer to -1 in the 8-cluster case. The same applies for emd- and rank emd correlation values, which are closer to 1 in this case. The correlation values for 16 clusters are obviously weaker than for 8 clusters. In both sections, the aggregate datapoints target is superior to the all datapoints target.
	
	\begin{table}[!ht]
		\tbl{Test error for ML models build alternatively with all- and aggregated datapoints.}
		{\begin{tabular}{ | c | c | c | c | c | c | }
				\hline
				{cluster size} & {datapoints source} & {datapoints target} & {test error} \\ \hline
				8	&	aggregate 	&	aggregate 	& 	\textbf{20.19} 	\\ \hline
				8	&	all 		&	aggregate 	& 	21.37	\\ \hline \hline
				8	&	aggregate 	&	all 		& 	\textbf{26.25}	\\ \hline
				8	&	all			& 	all 		&	26.68	\\ \hline \hline
				16	&	aggregate	& 	aggregate 	&	\textbf{20.47}	\\ \hline
				16	&	all			& 	aggregate 	& 	21.32	\\ \hline \hline
				16	&	aggregate	& 	all 		& 	\textbf{25.97}	\\ \hline
				16	&	all			&	all 		&	26.52	\\ \hline \hline
		\end{tabular}}
		\label{tab:train_test_errors}
		\begin{tabnote}
			Test errors for the same number of clusters can be accurately compared when the \textit{datapoints target} is the same. All models above were build using extreme gradient boosting.
		\end{tabnote}
	\end{table}
		
	\begin{table}[!ht]
		\tbl{Resulting correlation values for ML models built using all- and aggregated datapoints.}
		{\begin{tabular}{ | c | c | c | c | c | c | c | }
				\hline
				{cluster size} & {datapoints source} & {datapoints target} & cosine & rank\_cosine & emd & rank\_emd \\ \hline
				8	&	aggregate 	&	aggregate 	& 	\textbf{-0.53}	&	\textbf{-0.52}	&	0.30	&	0.17 	\\ \hline
				8	&	all 		&	aggregate 	& 	-0.53	&	-0.43	&	\textbf{0.37}	&	\textbf{0.27}	\\ \hline \hline
				8	&	aggregate 	&	all 		& 	-0.35	&	-0.36	&	0.20	&	0.14	\\ \hline
				8	&	all			& 	all 		&	\textbf{-0.41}	&	\textbf{-0.43}	&	\textbf{0.34}	&	\textbf{0.25}	\\ \hline \hline
				16	&	aggregate	& 	aggregate 	&	-0.16	&	-0.11	&	0.10	&	0.05	\\ \hline
				16	&	all			& 	aggregate 	& 	\textbf{-0.18}	&	\textbf{-0.17}	&	\textbf{0.22}	&	\textbf{0.17}	\\ \hline \hline
				16	&	aggregate	& 	all 		& 	-0.09	&	-0.06	&	0.08	&	0.00	\\ \hline
				16	&	all			&	all 		&	\textbf{-0.10}	&	\textbf{-0.11}	&	\textbf{0.17}	&	\textbf{0.08}	\\ \hline \hline
		\end{tabular}}
		\label{tab:correlation_values}
	\end{table}
						
	\subsection{Amenity area as similarity basis}
\begin{figure}[!ht]
		\centering
		\includegraphics[width=\textwidth]{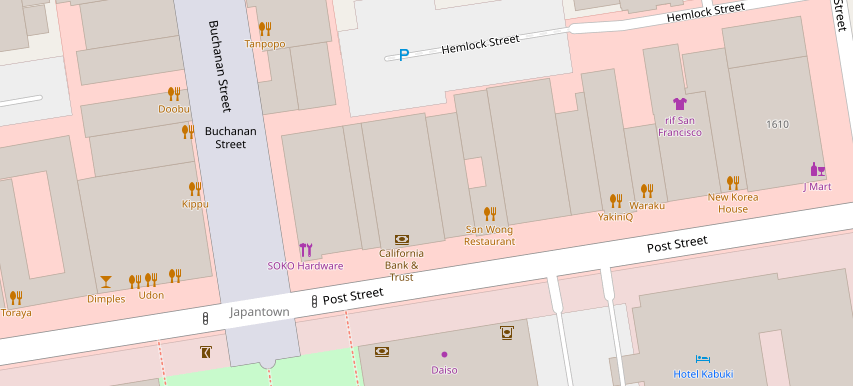}
		\caption{OSM screenshot emphasizing polygons as buildings and the amenities that are housed by them~\protect\cite{openstreetmap}.}
		\label{fig:amenity_polygons}
	\end{figure}

	We considered \textit{time spent} information to complement the amenity information as the basis for creating the mathematical representation. However, there are other factors that affect the parking demand towards an amenity.
	Obviously, one of them is the amount of people visiting the amenity.
	As we do not have data about this aspect, we use the area of the amenity as a proxy, assuming that larger places would have more visitors.
	OpenStreetMap provides a polygon layer for a certain geographic bounding box, which contains information across all the surfaces in that region. See \cref{fig:amenity_polygons} for a visualization of the polygons and their areas in OSM. 
	
	To extract this information we investigated two options.
	\begin{romanlist}
		\item \textbf{Polygon containing POI}. 
		Matching the amenities' POIs with the containing OSM polygon and then computing the polygon areas per amenity was the option tried first. 
		This has several drawbacks. 
		The relation POI $\colon$ polygon is in practice by no means 1 $\colon$ 1.
		Many cases arose where multiple POIs were contained by the same polygon, in which situation the area was split between them; a POI might also be on the edge of several polygons, in which case we have to either (arbitrarily) assign it to the first polygon or to all. 
		The deciding factor against this approach was, however, the fact that the \textit{coefficient of variation}, i.e., the ratio between the standard deviation and the mean of the sample, is larger than $1$, i.e., $2.1$ to be precise.  
		
		\item \textbf{Amenity attribute in polygon layer}. The other option was to use the \textit{amenity} attribute from the polygon layer of the region. 
		We could avoid the cumbersome matching by leveraging solely the polygon layer and calculating the amenity area mean and its standard deviation. The results are listed in \cref{tab:amenity_area_values}. 
		On top of that, the coefficient of variation is $0.9$ in this case, significantly lower than before. 
		
		Note that we have reduced the values in the table by a factor of 20, as it turned out that the actual mean and standard deviation were large enough to make the emd Gaussian computation extremely slow. As the standard deviation is linear with regard to the mean, both mean and standard deviation values were reduced conveniently. 
		
	\end{romanlist}
	
	\begin{table}[!ht]
		\tbl{Amenity area values gathered and averaged from OMS polygon layer for the SF\textit{park} region.}
		{\begin{tabular}{ | l | c | c | c || l | c | c | c |}
				\hline		
				\textbf{amenity name} & \textbf{mean} & \textbf{stdev} & \textbf{cat} & \textbf{amenity name} & \textbf{mean} & \textbf{stdev} & \textbf{cat} \\ \hline
				arts\_centre & 68 & 60 & 2 & bank & 39 & 20 & 2 \\ \hline
				bar & 19 & 8 & 1 & bicycle\_parking & 8 & 7 & 1 \\ \hline
				biergarten & 11 & 12 & 1 & brokerage & 39 & 9 & 2 \\ \hline
				bus\_station & 588 & 737 & 3 & cafe & 17 & 10 & 1 \\ \hline
				car\_rental & 70 & 43 & 2 & car\_wash & 43 & 48 & 2 \\ \hline
				childcare & 101 & 130 & 3 & cinema & 75 & 43 & 2 \\ \hline
				clinic & 61 & 32 & 2 & community\_centre & 52 & 74 & 2 \\ \hline
				conference\_centre & 401 & 519 & 3 & courthouse & 459 & 201 & 3 \\ \hline
				dentist & 17 & 12 & 1 & doctors & 324 & 568 & 3 \\ \hline
				embassy & 68 & 38 & 2 & fast\_food & 25 & 24 & 1 \\ \hline
				fire\_station & 52 & 27 & 2 & fountain & 24 & 22 & 1 \\ \hline
				fuel & 25 & 27 & 1 & library & 102 & 124 & 3 \\ \hline
				marketplace & 325 & 228 & 3 & music\_rehearsal\_place & 33 & 15 & 1 \\ \hline
				nightclub & 32 & 9 & 1 & nursing\_home & 97 & 47 & 2 \\ \hline
				parking & 182 & 309 & 3 & pharmacy & 65 & 38 & 2 \\ \hline
				place\_of\_worship & 60 & 62 & 2 & police & 137 & 124 & 3 \\ \hline
				post\_office & 39 & 11 & 2 & pub & 25 & 25 & 1 \\ \hline
				public\_building & 280 & 236 & 3 & recycling & 28 & 20 & 1 \\ \hline
				restaurant & 22 & 16 & 1 & school & 740 & 1280 & 3 \\ \hline
				social\_centre & 30 & 21 & 1 & social\_facility & 356 & 801 & 3 \\ \hline
				stripclub & 50 & 10 & 2 & studio & 268 & 307 & 3 \\ \hline
				swimming\_pool & 16 & 9 & 1 & swingerclub & 27 & 4 & 1 \\ \hline
				theatre & 174 & 191 & 3 & toilets & 7 & 5 & 1 \\ \hline
				training & 72 & 94 & 2 & veterinary & 21 & 7 & 1 \\ \hline
		\end{tabular}}
		\label{tab:amenity_area_values}
		\begin{tabnote}
			The mean and standard deviation values were reduced by a 20x factor. The categories for \textit{cosine vectors} are 0 - 35, 36 - 100, 100+.
		\end{tabnote}
	\end{table}
	
	By applying the procedure using the amenity area as similarity basis, we obtain the correlations values listed in \cref{tab:correlation_amenity_area} for 8- and 16 cluster configurations. The superiority of the correlation values using \textit{time spent} value is observed both for cosine- and rank cosine correlations, which are closer to -1, and for the emd- and rank emd correlations respectively, which are nearer to 1. All correlation values for 16 clusters are, however, relatively weak in absolute measures.
	
	\begin{table}[!ht]
		\tbl{The correlation results computed using similarity values based on \textit{amenity area}. }
		{\begin{tabular}{ | c | c | c | c | c | c | c | }
				\hline
				{cluster size} 	& {amenity type} 	& datapoints (train/test) 	& cosine 	& rank\_cosine & emd & rank\_emd \\ \hline
				8 				& {time spent} 		& agg/agg 		& \textbf{-0.53}	& \textbf{-0.52}		&	\textbf{0.30}	&	\textbf{0.17} \\ \hline
				8 				& area 				& agg/agg 		& 0.05	&	0.11	&	0.14	&	0.22 \\ \hline \hline
				16 				& {time spent} 		& agg/agg 		& -0.16	&	-0.11	&	0.10	&	0.05 \\ \hline
				16 				& area 				& agg/agg 		& -0.09	&	-0.03	&	0.09	&	0.07 \\ \hline
		\end{tabular}}
		\begin{tabnote}
			All models above were build, trained, and tested on aggregated datapoints.  
		\end{tabnote}
		\label{tab:correlation_amenity_area}
	\end{table}
							
	\subsection{Extended prediction models}
	An alternative to building mathematical representation of city areas and applying similarity functions is to let the machine learning model find out the similarities by itself. 
	One can choose to add the city data as further training information for clusters. The purpose is to produce better predictions by leveraging unknown patterns in the city data. Hence, features representing the \textit{cosine} and \textit{emd} functions are added to the model, as follows: 
	\begin{romanlist}
		\item k features corresponding to the k categories the amenities are split in, i.e. the magnitudes of the vectors for each category;
		\item a feature corresponding to the emd Gaussian value for that cluster, loosely interpreted as the "total accumulated time spent value" for that cluster, in case of \textit{time spent}, or the "total accumulated area" among all amenities in that cluster, for the amenity area; mathematically it is expressed in \cref{eq:emd_gaussian_magnitude}
		\begin{equation}
		feature(emd)=x \cdot f(x)
		\label{eq:emd_gaussian_magnitude}
		\end{equation}
		{\centering
			where $x$ = 
			$
			\begin{dcases}
			\text{duration in minutes}, & \text{if sim = \textit{time spent}} \\
			\text{area in square meters}, & \text{if sim = \textit{amenity area}}.
			\end{dcases}
			$ \\
			
			and $f$ is the constructed emd Gaussian function that equals the number of amenities in the cluster for any value x.}
	\end{romanlist}
	
	We subsequently build \textit{extended machine learning models} that additionally contain the features above. Since these features are identical for all datapoints in a cluster, the model needs to be trained on multiple clusters. Therefore, models on $n-1$ out of $n$ clusters will be build and tested on the remaining cluster. These \textit{all-but-one} or \textit{total models} will be constructed for all $n$ combinations of $n-1$ clusters and their averaged test errors will compared to total models that contain the normal features. To build the models, the same methods and parameters are used as described before.
	
	The resulting test errors are shown in \cref{tab:extended_models_comparison} for various number of clusters. In three out of four cases, the addition of the two features did not help with finding better parking occupancy values for the test clusters, at least using the gradient boost model we used in our experiments. For the 16-cluster case, the total model returns a superior result. In the one case, the simple total models achieved a better prediction performance. 
	
	\begin{table}[!ht]
		\tbl{Total models extended with cosine and emd features compared to total models with the previous feature set.}
		{\begin{tabular}{ | c | c | c | c | }
				\hline		
				\textbf{cluster size}  & \textbf{model} & \textbf{test error average} \\ \hline
				4 & {xgb} & \textbf{14.92} \\ \hline
				4 & {xgb total} & 16.18 \\ \hline \hline
				8 & {xgb} & \textbf{18.12} \\ \hline
				8 & {xgb total} & 19.58 \\ \hline \hline
				16 & {xgb} & 18.19 \\ \hline
				16 & {xgb total} & \textbf{18.09} \\ \hline \hline
				32 & {xgb} & \textbf{18.65} \\ \hline
				32 & {xgb total} & 20.38 \\ \hline
		\end{tabular}}
		\label{tab:extended_models_comparison}
	\end{table}
	
	\section{Further Possible Variations}
	To further investigate parking occupancy prediction given the assumptions in this work, there are several improvements or alternative approaches that can be experimented with:
	
	\begin{romanlist}
		\item \textbf{Use parking data from other locations.}
		In the present work, several pieces of data could not be integrated because of merging issues, i.e., the location units among multiple datasets did not coincide.
		Traffic, events, weather, etc. might improve estimation results and hence the final estimations for clusters without parking data.
		Other sources for parking occupancy data can be found for the cities of Cologne~\cite{cologne_data}, Zurich~\cite{zurich_data}, Santa Monica~\cite{santa_monica_data}.
		In Germany, Deutsche Bahn provides an API to obtain data from parking around train stations~\cite{dbbahn_data}.
		Data pertaining to street occupancy is, however, hard to find.
		At the time of writing, open data portals mostly provide the location of parking lots, parking meters, parking prices and opening times, if applicable.  
		
		\item \textbf{Gather better \textit{time spent} information}
		Accessing data on the time durations people spend in various amenities is restricted by Google by not providing an API for it. Other social media services, such as Foursquare offers information on how busy an amenity is by means of the number of check-ins that people send from their phones\footnote{https://developer.foursquare.com/docs/api/endpoints}  \footnote{https://developer.foursquare.com/docs/api/venues/details}. The API is however subject to costs and we have not tried it. By gathering more information automatically than we managed to collect manually, the evaluating similarity values based on the \textit{time spent} information will likely increase in accuracy and so will the resulting correlation values.  
		
		\item \textbf{Use more OSM (meta)data}.
		The mathematical representations in the present work are relying on public amenities available from OpenStreetMap.
		OSM has great potential as a collaborative map service but it currently still lacks many pieces of information that could be useful (in comparison with Google's \textit{time spent} information, for example).
		Data such as opening hours, if it would be widely available, would be interesting to include in the mathematical representations, which would then take into account the number of public amenities that are available at a certain point in time.
		Overall, more and finer city data, together with an appropriate representation and similarity function could eventually improve the occupancy estimations for clusters without parking data. 
		
		\item \textbf{Experiment with other clustering methods}
		Our approach is dependent on the fact that the computed K-Means clusters are of approximately the same size. Ideally, this would mean that the number of amenities of a certain \textit{time spent} value is virtually equal between clusters. But since for both \textit{cosine similarity} and emd it is not the absolute number of amenities for a certain \textit{time spent} value that matters, instead the relative number between certain \textit{time spent} values, it may be that clusters of clearly opposite sizes are very similar. All the more is the reason to experiment with other clustering methods, such as DBSCAN and OPTICS. Both focus on detecting clusters based on density of neighborhoods, which may translate in practice into separating regions in the city that are more sparse, e.g., large office areas, from regions that are very dense, e.g., residential or old town areas.
		
		\item \textbf{Apply semi-supervised machine learning}.
		Another relevant machine learning approach in this case is based on organizing the city areas as an undirected graph.
		The vertices represent the clusters with their respective occupancy data, while the edges between them are assigned similarity values.
		Initially, only a part of the vertices have the occupancy value known, i.e., the clusters with parking data, while the rest has undetermined occupancy, i.e., the clusters without parking data.
		At each step, the value for a vertex whose value is undetermined is being computed by considering the occupancies of the linked vertices and their corresponding similarity values.
		
		\item \textbf{Accounting for completely filled parking spaces in the vicinity}.
		Currently, we only take information from nearby amenities into account for the model.
		However, one could imagine that when nearby areas have completely full parking areas, then cars will `spill over' to the area under investigation.
		A model having to account for this type of factors would, however, be a lot more complicated to express.
	\end{romanlist}

	\section{Conclusion}
	We presented our work on approximating street parking occupancy in cities.
	Given the fact that parking sensors cannot capture all parking areas and assuming that parking demand can be modeled using urban features, we proposed an alternative solution to the ones previously developed for this problem.
	The data used was composed of parking data from the SFpark project, that investigated parking pricing in San Francisco, and OpenStreetMap. The two data sources were merged and afterwards split into small city areas that separated the regions that contained parking data and other regions that did not contain parking data.
	We built mathematical representations for amenities, any form of facility or building in a city, and their \textit{time spent} information, visiting duration that have been gathered using smartphones. This led to computing similarity values between city areas by applying functions such as \textit{cosine similarity} and \textit{earth mover's distance} to the mathematical objects.
	The K-Means algorithm was used to cluster the city areas, while four methods were employed to train models for the clusters: \textit{decision trees}, \textit{support vector machines}, \textit{multilayer perceptrons} and \textit{extreme gradient boosting}.
	The occupancy estimations for clusters \textit{without} parking data were defined in terms of model estimations from clusters \textit{with} parking data and the corresponding cluster similarity values.
	The estimations are expressed as intervals which extend the model prediction values by the magnitude of the similarity values. 
	
	The data source for our work, the SF\textit{park} project, gathered parking data for more than 2 years starting in 2011 in San Francisco and now offers it for free usage.
	The city data was collected from OpenStreetMap as amenity information, and from Google Places as stay duration values.
	Both sources are open and free of charge.
	Over 30 types of public amenities were found in the San Francisco blocks and data from over 470 Google Places sources was collected.
	
	Following our experiments, the best machine learning model for our problem setting turned out to be extreme gradient boosting.
	We used the clusters \textit{with} parking data for the evaluation of the similarity values and calculated correlation coefficients between the similarity values and the estimation errors, using both absolute values and ranks.
	The best correlation were reached for the 100m \textit{merge distance} for 8 clusters.
	In the same configuration, both \textit{cosine similarity} and emd reached their best results from all the experimented configurations.
	
	We further investigated the test error and correlation values when models were built and tested on all datapoints instead of aggregated datapoints. The aggregation factor has been shown to play a role in finding the best estimations.
	
	Overall, we touched upon several aspects that influence parking occupancy and provide a ground for further experimentation, which can further improve parking occupancy estimations when transferring models from monitored to unmonitored regions.

\bibliographystyle{ws-ijait}
\bibliography{parking_paper_references_ijait}

\end{document}